\documentclass[letterpaper, 10 pt, conference]{ieeeconf}  
\usepackage{graphicx}
\usepackage{amsmath,amssymb} 
\usepackage{color}
\usepackage{caption}
\usepackage{hyperref}
\hypersetup{
    colorlinks=true,
    linkcolor=blue,
    filecolor=magenta,      
}

\IEEEoverridecommandlockouts                              

\title{\LARGE \bf
Visual Representations for Semantic Target Driven Navigation
}

\author{
Arsalan Mousavian$^{2,\circ}$, Alexander Toshev$^{1,\star}$, Marek Fi\v{s}er$^{1}$, Jana Ko\v{s}eck\'a$^{3,\circ}$, Ayzaan Wahid$^{1}$ James Davidson$^{4}$ \\
$^\circ$Work done while employeed at Google, $^{1}$Robotics at Google, \\$^{2}$NVIDIA, $^{3}$George Mason University, $^{4}$Third Wave Automation
\thanks{$^\star$Primary contact: Alexander Toshev, toshev@google.com}
}

\begin{document}

\maketitle
\thispagestyle{empty}
\pagestyle{empty}

\begin{abstract}
What is a good visual representation for navigation? We study this question in the context of semantic visual navigation, which is the problem of a robot finding its way through a previously unseen environment to a target object, e.g.~{\em go to the refrigerator}.  Instead of acquiring a metric semantic map of an environment and using planning for navigation, our approach learns navigation policies on top of representations that capture spatial layout and semantic contextual cues. 
 
We propose to use semantic segmentation and detection masks as observations obtained by state-of-the-art computer vision algorithms and use a deep network to learn the navigation policy. The availability of equitable representations in simulated environments enables joint training using real and simulated data and alleviates
the need for domain adaptation or domain randomization commonly used to tackle the sim-to-real transfer of the learned policies. 
Both the representation and the navigation policy can be readily applied to real non-synthetic environments as demonstrated on the Active Vision Dataset~\cite{active-vision-dataset2017}. Our approach successfully gets to the target in $54\%$ of the cases in unexplored environments, compared to $46\%$ for a non-learning based approach, and $28\%$ for a learning-based baseline.

\end{abstract}

\section{Introduction}

Visual perception is one of the key capabilities of intelligent robotic agents, enabling them to purposefully act in an environment. The question then naturally arises as to  what is the most appropriate representation derived from visual observations that can support various robotic tasks? We study this question in the context of target-driven semantic visual navigation, where an agent deployed in an unexplored environment is tasked to navigate to a semantically specified goal, e.g. {\em go to refrigerator}. 

Earlier work on navigation has been fragmented and has rarely resulted in reliable systems that can be comprehensively evaluated in a variety of environments. Traditional approaches focus on 3D metric and semantic mapping of the environment~\cite{SLAM16,armeni_cvpr16} followed by path planning and control. Such approaches typically require building a 3D map ahead of time, and reliable localization for mapping and path following. They typically do not exploit general semantics and contextual cues in the decision making stage. 

More recently, the success of data driven machine learning strategies for a variety of control and perception problems opens new avenues for overcoming the limitations of previous approaches~\cite{deepMindRL_ICLR2017,ControlMemory,OneShotLearning,TargetDriven16,House3D_NIPS17}. The gist of these methods is to directly learn a mapping between raw observations and actions in an end-to-end fashion for the task. 
The additional appeal of learning based approaches is the capability of leveraging previous navigation experiences in novel similar environments, with or without a map.  

\begin{figure}[t]
\centering
\begin{tabular}{@{\hskip3pt}c@{\hskip3pt}c@{\hskip3pt}c}
  \includegraphics[width=0.15\textwidth, height=0.12\textwidth ]{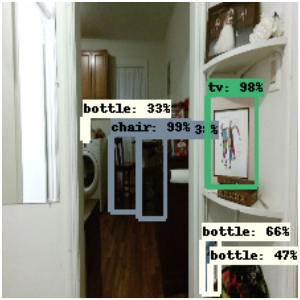} &
  \includegraphics[width=0.15\textwidth, height=0.12\textwidth ]{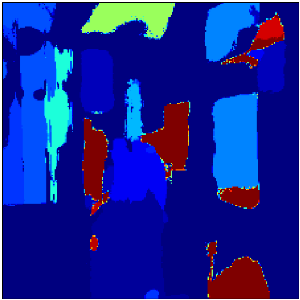} &
  \includegraphics[width=0.15\textwidth, height=0.12\textwidth ]{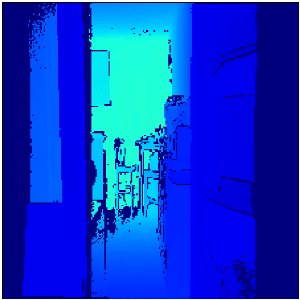} \\
  
  \includegraphics[width=0.15\textwidth, height=0.12\textwidth ]{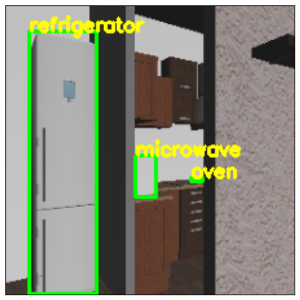} &
  \includegraphics[width=0.15\textwidth, height=0.12\textwidth ]{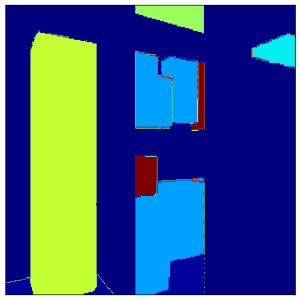} &
  \includegraphics[width=0.15\textwidth, height=0.12\textwidth ]{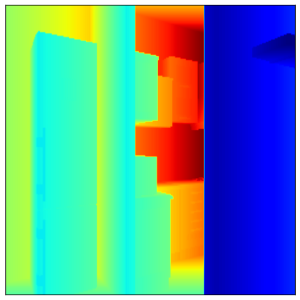} \\
\end{tabular}
\caption{Visualization of detections (left), segmentations (middle) and depth (right). While for real data (top) we use off-the-shelf detector and segmenter, the output on the simulated data (bottom) comes from the labels in this data and the renderer.}
\label{fig:sseg_objdet_viz}
\end{figure}
In this work we propose to use high-level semantic and contextual features included in segmentation and detection masks and learn a navigation policy from these observations.  
Given extensive navigation experience in similar environments during training, the agent can discover commonly encountered objects and contextual cues and learn policies that can generalize to previously unseen environments. 
We demonstrate that the proposed visual representations and associated policy enables {\bf better generalization} of the navigation model trained on the smaller dataset. Furthermore, these {\bf transferable representations} enable simultaneous use of real and simulated environments for training, without the need for visual domain adaptation or randomization commonly used to transfer between simulated and real environments. These segmentation and detection masks capture the outlines of a wide range of foreground and background objects, and as such provide a detailed description of the scene. 
We perform a thorough investigation of the above representations and combine and contrast them with raw RGB and depth inputs. 

A further contribution of this work is the choice of the model and training of the navigation policy. At training time, we use an optimal path planning algorithm as a form of stronger supervision to estimate the progress toward the goal after taking an action.
The navigation policy model learns to predict this progress value and uses it to select an action at test time. We investigate several models -- a feed-forward memory-free model as well as a model with internal state implemented using an LSTM~\cite{hochreiter1997long}.

We use both synthetic~\cite{song2016ssc} and real environments ~\cite{active-vision-dataset2017} to train and evaluate the learned navigation policy with a detailed ablation study of the choice of visual representations and model architecture. This proposed model reaches the target successfully 54\% of the time on previously unseen environments.
Code and visualization of the agent's paths have been made available at \url{https://github.com/tensorflow/models/tree/master/research/cognitive_planning}.

\section{Related work}
The proposed work is related to a range of different approaches towards visual navigation. Classical techniques for navigation typically start with map construction, followed by planning and execution of the planned trajectories. The mapping stage uses visual observations for simultaneous localization and mapping (for an extensive review see ~\cite{SLAM16,armeni_cvpr16}). Given the model of the environment, state-of-the art planning and control strategies can be employed~\cite{prm-rl-ICRA18}.

More recent methods bypass the mapping and explicit pose estimation steps and learn navigation strategies directly. The existing learning based approaches differ in the architecture, size and complexity of the models, cost of the training stage, generalization capability, and type of supervision. The existing formulations consider either a {\em point-to-point} navigation strategy, where the goal is given as a coordinate in an ego-centric coordinate frame~\cite{CognitiveMapping} or a {\em target-driven} navigation strategy, where the image of the target is given as an input~\cite{TargetDriven16}.

Reinforcement learning (RL) strategies and associated deep learning architectures are discussed by Mirowski et al.~\cite{deepMindRL_ICLR2017,mirowski2018nips}. The experiments are carried out in synthetic 3D maze environments with a single goal. The observations in these environments are simplistic mazes and do not exhibit the complexity of real world settings. Mirowski et al.~\cite{deepMindRL_ICLR2017} suggest overcoming the difficulties associated with sparse rewards by using auxiliary losses for depth prediction and loop closure detection. For simulated indoor environments, basic RL strategies (feed-forward A3C, A3C with LSTM and Direct future prediction method by Dosovitskiy et al.~\cite{Dosovitskiy_ICLR2017}) for point goal and room goal prediction are benchmarked in the Minos environment~\cite{MINOSFall17}. 

For the related problems of object reaching with a manipulator and quadcopter flying, Sadeghi et al.~\cite{SadeghiCVPR18,SadeghiRSS17} use synthetic data with domain randomization to learn control strategies.

Gupta et al.~\cite{CognitiveMapping} and Khan et al.~\cite{MemoryControl17} use value iteration networks~\cite{VIN} and imitation learning to learn navigation strategies. Bruce at al.~\cite{bruce2018corl} learn navigation policies from a single traversal of the environment. 


The broader area of active and embodied perception has received increased interest focusing on navigation, task planning, and visual question answering. These tasks have motivated different simulation environments derived from SUNCG~\cite{song2016ssc} or Matterport3D~\cite{Gibson_CVPR18,MINOSFall17} datasets. While there are numerous architectures for visual question answering~\cite{IQAGordon,EmbodiedQA,House3D_NIPS17}, with the exception of~\cite{House3D_NIPS17}, the evaluation metrics do not explicitly consider the navigation component. 

%
%

\section{Navigation Model}
\subsection{Setup}\label{sec:setup}
We address the problem of navigating to a target, defined by its class label, using purely visual observations. 
This problem can be represented as a partially observable Markov decision process (POMDP) $(S, A, O, P, R_c)$. The state space $S$ consists of the pose of an agent, which is not observable. The action space $A$ is a discrete set of turns and translations of pre-defined lengths inducing a lattice structure-- described in detail in the experiments section. 
Observation space $O$ are raw RGB and depth images. The probability $P(s'|s,a)$ reflects the transition to a new state $s'$ after an execution of an action $a$ in the current state $s$. Finally, the reward $R_c$ expresses the distance of the agent from the target $c$ and can be defined as the negative of the shortest path between the current state $s\in S$ and the target $c$:~$R_c=-d(s, c)$.

Within the above setup, our navigation policy is represented by a Neural Network $\pi(a|o;c)$ which given the target class label $c$ and an observation $o\in O$ predicts an action $a\in A$. The policy generates a sequence of actions that move the agent from its starting position to the target.

\begin{figure*}
    \centering
    \includegraphics[width=0.9\textwidth]{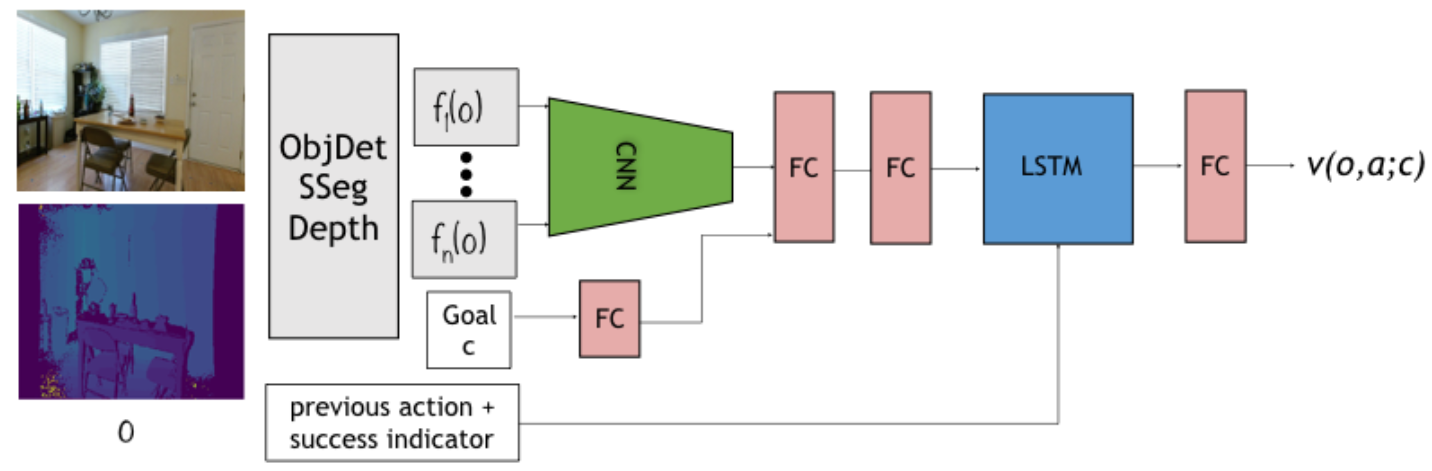}
    \caption{Model Overview: RNN takes concatenation of joint representation of observations and target in addition to previous action and success indicator. The model predicts the cost $v(o,a; c)$ of taking action $a$ at the current state. Finally, the controller takes the action with the lowest cost (see.~(\ref{eq:controller})).}
    \label{fig:model_overview}
\end{figure*}

\subsection{Visual Representations}
The focus of this work is representations that can be extracted from the visual observations. Such representations, denoted by $f(o)=(\ldots, f_i(o), \ldots)$, are the bridge between the raw signal and the controller. Thus, the model can be written as $\pi(a|f(o); c)$.
In robotic applications the raw observations consist of an RGB image, which is often augmented with a depth image, derived from Radar, LiDAR or stereo. Therefore, a common choice of $f$ is a ConvNet~\cite{szegedy2017inception,he2016deep}, which is typically designed for and pre-trained on a large classification dataset~\cite{deng2009imagenet}. Such a network is used to produce a vectorial embedding of the image.

To this end we consider the following representations:

\textbf{\textit{Det}}: A set of filled object detection masks. We use the Faster R-CNN~\cite{fasterRCNN_2015} object detector trained on COCO~\cite{linCOCO2014}. The detector output is converted to a $H \times W \times C_{det}$ mask where $H$ and $W$ are the height and width of the image and $C_{det}$ is the total number of object categories in COCO. The $j^\textrm{th}$ channel contains all detected boxes for the $j^\textrm{th}$ class as 0/1 masks.

\textbf{\textit{SSeg}}: The output of the segmenter defined by Mousavian et al.~\cite{Mousavian3DV16} that is trained on NYU V2 dataset~\cite{NYUV2}. The resulting representation is a $H \times W \times C_{sseg}$ mask stack where $C_{sseg}$ is the number of categories in the NYU V2 dataset.

\textbf{\textit{Depth}}: The raw depth channel converted to $H \times W \times 2$ where the first channel is the normalized depth values coming from the sensor and the second channel is a binary mask that indicates whether a pixel has a valid or missing depth.

\textbf{\textit{RGB}}: The penultimate layer output of the ResNet-50~\cite{he2016deep} trained on ImageNet.

Furthermore, \textit{RGB} cannot be easily used on both real and simulated data as the network produces different outputs on the different domains. A large body of work focuses on domain adaptation~\cite{ganin2015unsupervised,bousmalis2017unsupervised,vir-to-real2018ijcai} and its applications to robotics \cite{graspGAN_2018,Gibson_CVPR18}. Such approaches are hard to train as they rely on still relatively unstable generative adversarial network setups.

Both \textit{Det} and \textit{SSeg} address some of the above challenges. They capture scene layouts, obstacles, and locations of target objects (see Fig.~\ref{fig:sseg_objdet_viz}). As we empirically show later, despite being lossy compared to \textit{RGB}, these representations capture most of the necessary information for navigation without the need for fine-tuning. Thus, large segmentation and detection networks are used without being part of the training of the  navigation controller, which makes the optimization of the latter easier and more stable.

Another advantage of \textit{Det} and \textit{SSeg} is that there is no need for domain adaptation. While semantic segmenation and detectors are used on real data, in simulation the object masks and bounding boxes can be generated by the renderer (see Fig.~\ref{fig:sseg_objdet_viz}). Then the gap between real and simulated data is related to the quality of the segmentation and detector. This setup is particularly timely as research on object detection~\cite{linCOCO2014}, semantic segmentation~\cite{DeepLab16} and depth estimation~\cite{Mousavian3DV16} has been propelled by deep learning methods with a variety of high performing models available. We show empirically that by using simulation with these representations we achieve substantial improvement without any domain adaptation. 

\subsection{Model}\label{sec:model}
The model $\pi(a|f(o);c)$ using the above representation is a Deep Neural Network (see Fig.~\ref{fig:model_overview}). In addition to the current observation the model takes as an input a description of the target in the form of a one-hot vector over a pre-defined set of classes. This secondary input modulates the network behavior.



The architecture of the network is shown in Fig~\ref{fig:model_overview}. The CNN architecture that we use for extracting the representation from each feature extractor $f_i(o)$ consists of three convolutional layers and a fully connected layer, all using ReLU as activation. The convolution layers have kernal sizes $8\times 8\times 8$, $4\times 4\times 16$, $3\times 3\times 16$ and  strides of $4,2,1$ respectively. The fully connected layer produces a $128$ dimensional embedding for each modality except \textit{RGB}, for which we use a  pretrained ResNet50. The input target $c$ is presented with a one-hot vector and gets projected to an embedding of size $128$. All the embeddings are concatenated and processed with a series of fully connected layers and an LSTM. The size of the fully connected layers and LSTM is 2048.

Furthermore, we experiment with explicitly using the previous action as an input to the model along with a binary indicator of whether it resulted in a collision. We assume that there is a collision detection module, which is commonly present in robot systems. This input helps the model choose a different recovery action when experiencing a collision.

\subsection{Training}
In our navigation setup we have full knowledge of the environment at train time. 
We consider a discrete setting where the environment is described by a graph $G_e$ with nodes representing a discrete set of states -- poses (locations and headings) of the agent. Edges in this graph correspond to possible state transitions.
Thus, for the problem of getting to an object, one can use a shortest path planning algorithm. At test time, however, the environment map (graph) is not known. Nevertheless, we would like to learn a controller which performs as close as possible to an optimal path by exploiting general contextual cues. Therefore, instead of employing Deep Reinforcement Learning, where at train time the agent is to discover the optimal path guided only by a reward definition, we use strong supervision from a path planning algorithm. 

The above setup falls into the domain of imitation learning. Common approaches include behavioral cloning \cite{argall2009survey} and Dagger \cite{ross2011reduction}. In such approaches the agent is trained to emulate an expert, in this case a path planner, producing demonstrations. One drawback of the above approaches in our setup, however, is that often times multiple actions at a given state can follow an optimal path. And when an action is not optimal, it isn't necessarily incorrect, as it can still lead the agent towards the goal, albeit not via the shortest path. Therefore, we train our agent to predict the cost of an action $a$, which is defined as the 'progress' toward the goal (an approximation of the negative of the value function in RL) -- the reduction of the shortest distance $d(s, c)$ from current state $s$ to the target $c$ after taking $a$:
\begin{equation}\label{eq:cost}
    y(s,a;c) = \sum_{s'} P(s'|s,a) d(s', c) - d(s, c)
\end{equation}
As defined above the value range of $y$ is $[-1, 1]$. To penalize collisions, we set $y(o,a;c)=+2$ if action $a$ leads to collision. Finally, if $a$ leads to the goal, we set $y(o,a;c)=-2$ to designate a 'stop' state. Otherwise, the value of the action 'stop' is set to +2 to discourage the agent to stop prematurely. The value of the above cost reflects the 'correctness' of an action.
Details of the transition model, $P(s'|s,a)$, are described in the experimental section. 

Using the above definition of state-action cost, we train a neural network $v(o, a; c)$ to predict the above cost directly from observations:
\begin{equation}
\label{eq:loss}
    Loss = \frac{1}{|C||S||A|}\sum_{c \in C}\sum_{s \in S}\sum_{a \in A} \left(v(o,a;c) - y(s,a;c) \right)^2
\end{equation}

The final navigation controller chooses at test time an action with the lowest predicted cost:
\begin{equation}\label{eq:controller}
    \pi(a|o;c) = \arg\min_{a\in A}v(o, a; c)
\end{equation}
Note that although we do not consider the history of observations explicitly in the above formulation, it is modeled implicitly by the LSTM component of our model. 

\noindent {\bf Discussion:} During training we use both real and simulated environments. The real data comes from dense scans of real houses. Both environments are discrete in nature -- we work with a pre-defined set of possible locations and orientations; the action space is discrete. Thus, in each state we can enumerate all possible next state-action pairs, as shown in the loss in Eq.~(\ref{eq:loss}). Contrary to most other POMDP setups, $S\times A$ is not prohibitively large and we are guaranteed to visit all state-action pairs eventually.  Thus, there is no need of state exploration guided by a policy.

\section{Experiments}\label{sec:experiments}

We use Active Vision Dataset (AVD)~\cite{active-vision-dataset2017} for evaluation. AVD consists of dense scans of 9 different homes, some of which have been scanned twice. In order to evaluate generalization capabilities, we use two different train/test splits. In the first split, which we refer to as \textit{similar environments}, the train and test environments are different scans of the same home (some of the objects are moved; scanning locations are changed). The second split, called \textit{different environments}, contains different homes for test and train. The houses which have been captured twice are indicated in the dataset.

As an additional source of training data, we use 
SunCG \cite{song2016ssc}, which is a large set of synthetically generated homes. We use a subset of 200 as defined in \cite{House3D_NIPS17}. All the evaluations are done on AVD because we want to see the effect of representation on the real data. 

 We consider a discrete set of actions \texttt{\{move\_forward, move\_back, move\_left, move\_right, stop, rotate\_ccw, rotate\_cw\}} and five semantic goal categories \texttt{\{dining\_table, refrigerator, television, couch, microwave\}}. The object detection categories are borrowed from COCO \cite{linCOCO2014} and the semantic labels are from NYU-v2  ~\cite{NYUV2}. We label the views that are closest to goal categories. We use a pre-trained object detector and semantic segmentation models for AVD and groundtruth detection and semantic segmentation masks for SunCG dataset. To match the settings of AVD, we use the same camera parameters, camera height, and action step size.

\subsection{Training Details}
Each minibatch consists of losses formulated over 8 random trajectories. Each trajectory is generated by selecting a random environment, from either AVD or SUNCG, generating a random goal and target and selecting the first 20 locations from this trajectory.
The output of all $f_i(o)$ except the raw image is resized to $64\times64$ pixels. Raw images are resized to $299\times299$ pixels to match the resolution at which the ResNet50 has been trained.
We use Adam Optimizer, learning rate of $10^{-4}$ and exponential decay schedule with decay rate of $0.98$ at every $1000$ steps. The model is optimized for $200\,000$ iterations on $40$ GPUs, which takes 16 hours. 

\subsection{Evaluation Settings}
We perform evaluations only on the test environments from AVD. The agent is run up to 100 steps, with action \texttt{stop} terminating the episode early. The agent performance is measured by \textit{success rate}-- defined as the portion of the runs in which the goal was reached. The agent is considered to have reached the goal if it is within 5 steps from any instance of the goal object (coming closer than that provides degenerate observations as the object often time covers the full FOV). Initial locations are chosen randomly and they are fixed between different experiments. Each initial location is evaluated for all the goal categories. 

\subsection{Analysis}
\subsubsection*{\textit{RGB} vs Semantic Representation}
\begin{figure}
      \includegraphics[width=0.5\textwidth,height=0.3\textwidth]{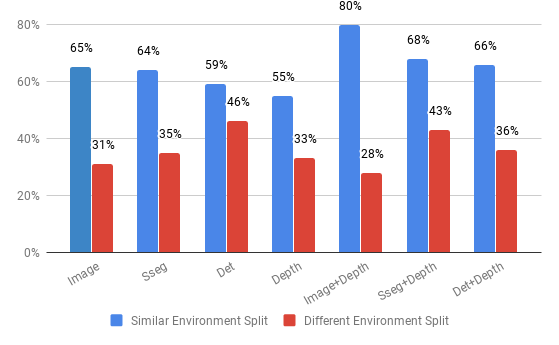} 

    \caption{Success rates of the models  on \textit{similar} (blue) and \textit{different environment split} (red) of AVD.}
    \label{fig:first_combo1}
\end{figure}

To contrast the proposed semantic representation with standard image embeddings, we use a ResNet50~\cite{he2016deep} to compute an image representation. This net is trained in an end-to-end fashion together with the policy. We show performance in Fig~\ref{fig:first_combo1}.

We observe that our representation generalizes better on unseen environments. This is shown by the higher performance on \textit{different env.}~split, where at test time we see totally different homes. This is true for both proposed semantic representations, \textit{SSeg} and \textit{Det}. Further, adding depth seems to hurt such generalization.

At the same time, a ResNet-based embedding allows the model to overfit on scans from the same homes. As such, this embedding might be more appropriate when we deploy the robot to environments we have seen during training.

\begin{figure}
      \includegraphics[width=0.5\textwidth]{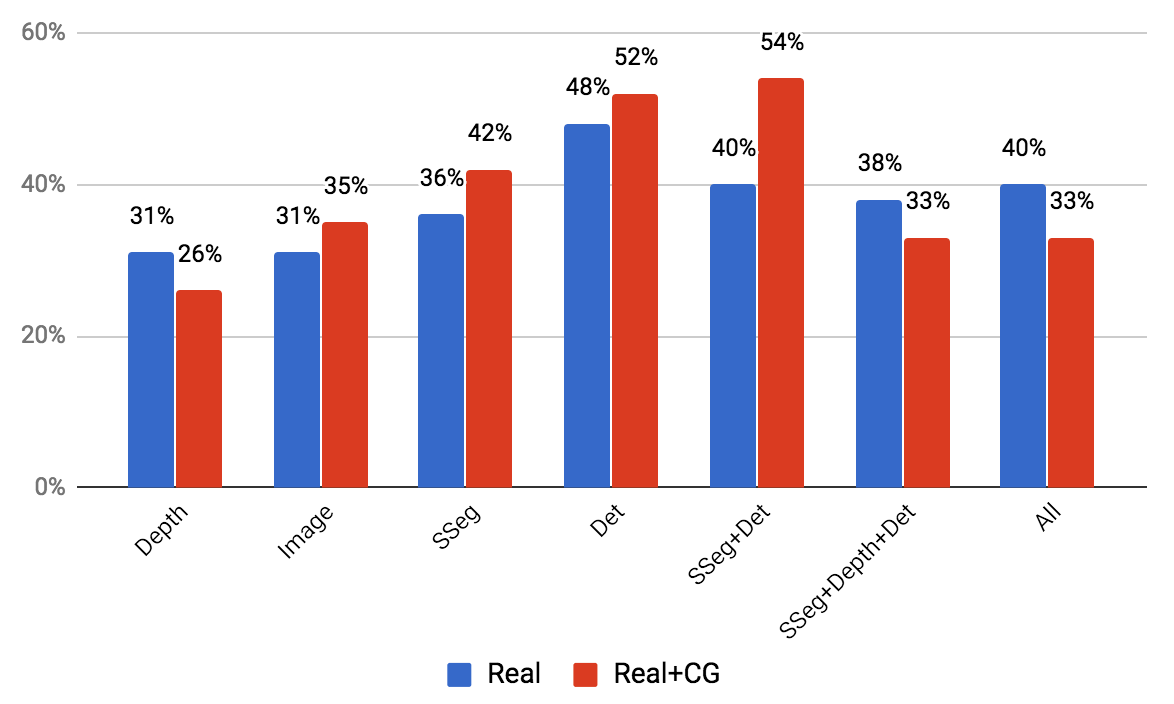}  \\
    
    \caption{Success rate of our model with different representations when trained using AVD only (blue) or AVD and SunCG (red).}
    \label{fig:first_combo2}
\end{figure}

\subsubsection*{Use of Synthetic Environments} In this section, we evaluate the effect of augmenting training with simulated data from SunCG dataset. During the training, an episode is sampled uniformly from AVD or SunCG. The evaluation is done on AVD only. 

Results are presented in Fig.~\ref{fig:first_combo2}. The use of simulated data in training improves the performance for \textit{Det} and \textit{SSeg} and their combination. This shows that the proposed representations are capable of using ample simulated data without the need for domain adaptation. The performance increase is in some cases over $10\%$. The reality gap, however, is an issue for raw observations, such as \textit{Depth}, where the performance drops. While depth is perfect on SunCG, on AVD depth is estimated using Kinect, which is noisy and has missing values. To mimic the imperfections of Kinect depth, depth values are clipped at 12m, perturbed by multiplying them with a scalar sampled uniformly from [0.9, 1.1], and $10\%$ of the depth measurements are removed at random.


    

\begin{figure}
\centering
  \includegraphics[width=0.4\textwidth]{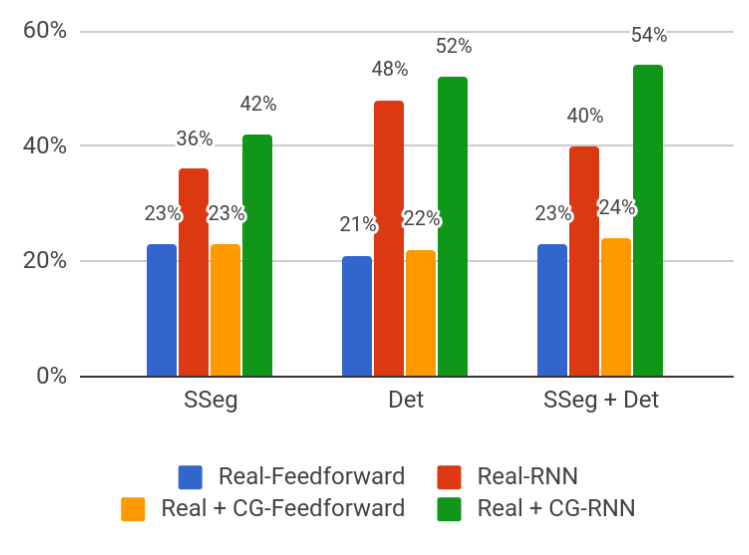} 

    \caption{Success rates of a feedforward vs recurrent model using AVD only (denoted by real) or AVD and SunCG.}.
    \label{fig:second_combo1}
    \vspace{-0.3cm}
\end{figure}

\begin{figure}
\centering
  \includegraphics[width=0.4\textwidth,height=0.34\textwidth]{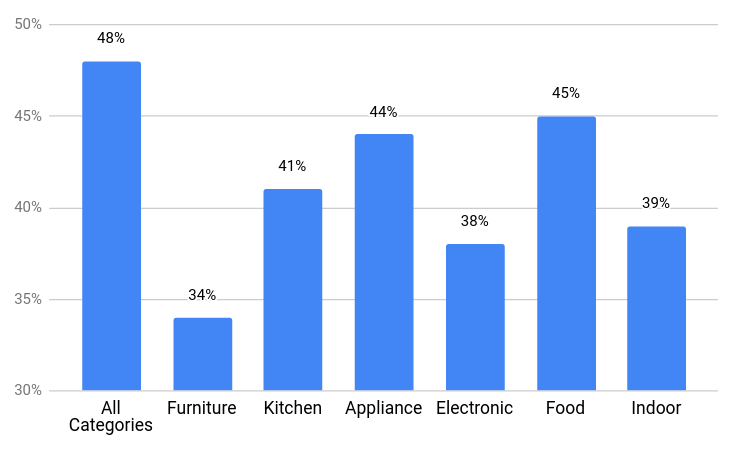} 
    
    \caption{Success rate of our model using all and a subset of the object classes from COCO in \textit{Det} representation.}
    \label{fig:second_combo2}
    \vspace{-0.3cm}
\end{figure}


\subsubsection*{Reactive vs Recurrent Model} To emphasize the importance of recurrence in our model, we present results of a feedforward network in Fig~\ref{fig:second_combo1}. We observe that the success rate drops significantly when the model does not have a state. Qualitatively, this is because a state-less model cannot correct past wrong actions, and often times keeps repeating them, leading to an oscillatory behavior (revisiting same state over and over again). Furthermore, we see that more training data does not rectify such behavior.

\subsubsection*{Importance of Different Object Classes} Our \textit{Det} representation is based on $80$ object types, as defined by COCO~\cite{linCOCO2014}. We would like to understand which of these classes are more important for the navigation agent. For this, we train several agents which use a subset of the all classes. In particular, in Fig~\ref{fig:second_combo2} we show performance of the model, where we remove one of the six superclasses \texttt{\{Furniture, Kitchen, Appliance, Electronic, Food, Indoor\}}.

The results show that the \texttt{Furniture} category, which contains large furniture items such as couches, tables, and beds, leads to the largest drop if removed. These objects are important because they are detectable from far away and exhibit strong correlation with the location of the agent in the house. On the other hand, \texttt{Food} bears little importance for the model, most likely due to its small size and lack of permanent locations.

\subsubsection*{Comparison with non-learning baseline} 
In this section, we compare the performance of our method with a semi-random search method. This baseline search strategy is not an optimal one. It is intended to give us an intuition for how a simple search strategy compares to our method. The baseline method, contrary to ours, is given access to the full graph of the environment $G_e$ and the pose for each view in the global coordinate system. The registered poses are provided in the dataset and are computed from a 3D reconstruction of all the images for each house.

The baseline agent has two modes. The agent enters the first mode if the object detector sees at least one instance of the target category $c$. In this mode, the location of the object is back-projected to the world coordinate system using the pose of the current view, depth channel, and the intrinsic parameters of the camera. The nearest view from $G_e$, that is directed toward the projected point cloud of the object, is chosen as a new destination. The shortest path is computed toward the designated vertex and the agent executes the shortest path without receiving any new observations. 

If the agent does not see the object, it takes a random action. Note that even if the object is detected from very far away, the shortest path algorithm on $G_e$ avoids obstacles and computes a path for the part of the scene that is not explored by the agent. 

\begin{figure}[h]
    \centering
    \label{fig:comparison}
    \begin{tabular}{cc}
         \includegraphics[width=0.22\textwidth,height=0.2\textwidth]{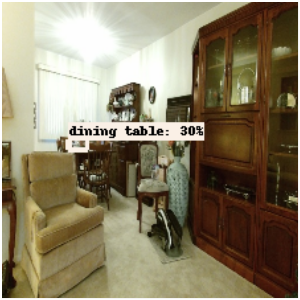} & \includegraphics[width=0.22\textwidth,height=0.2\textwidth]{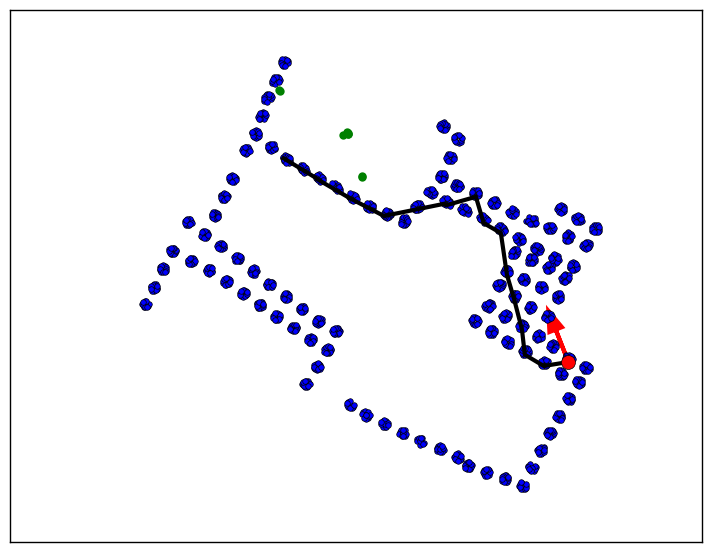} \\
         \includegraphics[width=0.22\textwidth,height=0.2\textwidth]{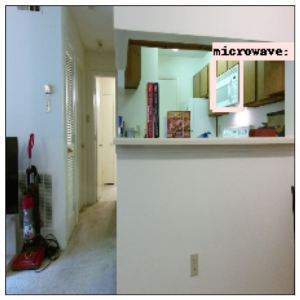} & \includegraphics[width=0.22\textwidth,height=0.2\textwidth]{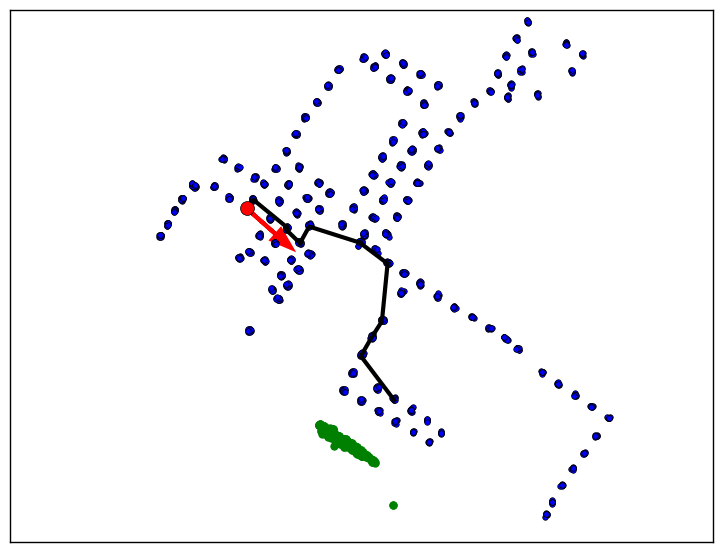} \\
    \end{tabular}
    \caption{Each row demonstrates the first mode of the baseline method. After the target object has been seen and detecter (left), the baseline uses depth and a pre-specified map (right), to backproject the detection onto the map (object displayed in green) and compute a path from its current location (red arrow) to the detection.}
    \label{fig:random_baseline}
    \vspace{-0.3cm}
\end{figure}

Fig~\ref{fig:random_baseline} illustrates an example of such a situation. Note that the accuracy of the baseline methods is affected 
by the performance of the object detector. 
The baseline method achieves $46\%$ success rate over the same evaluation set with the same initial poses. This shows the strength of our method, which can achieve superior performance with significantly less knowledge of the environment. 

%
%

\begin{figure}[h!]
    \centering
    \begin{tabular}{cc}
         \fbox{\includegraphics[width=0.2\textwidth]{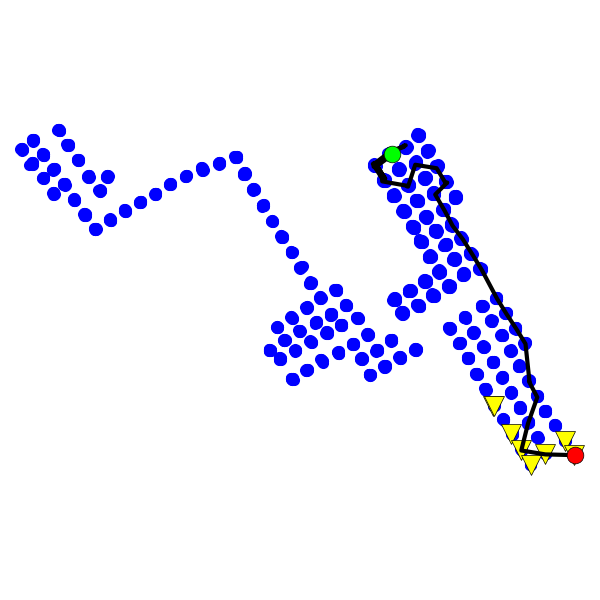}} & \fbox{\includegraphics[width=0.2\textwidth]{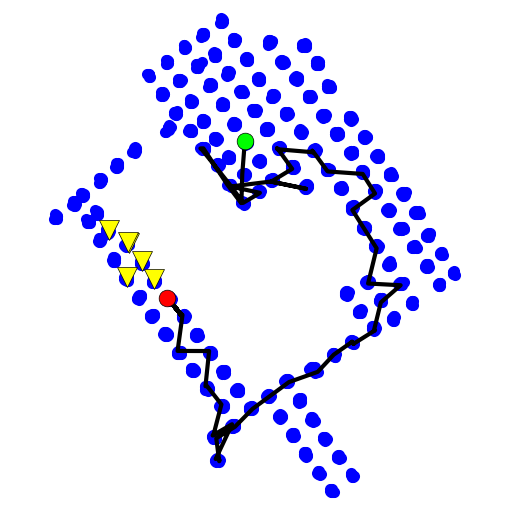}} \\
         Target: Couch & Target: Fridge \\
         \fbox{\includegraphics[width=0.2\textwidth]{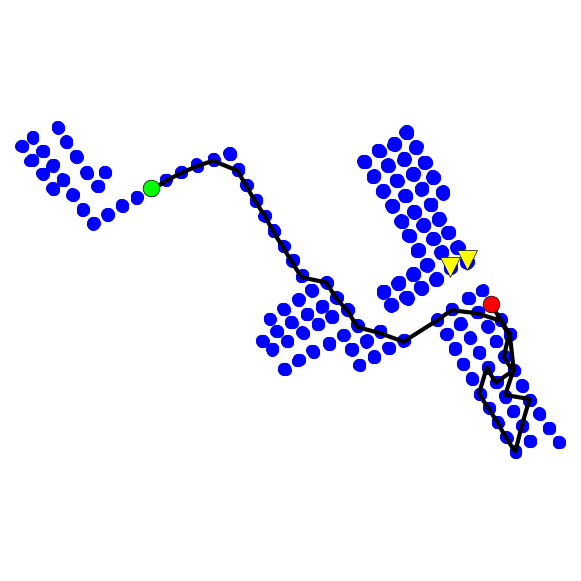}} &
         \fbox{\includegraphics[width=0.2\textwidth]{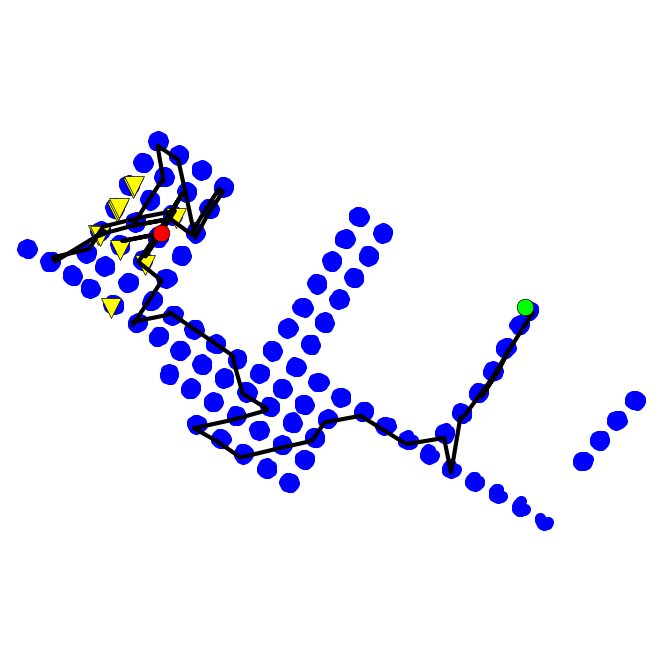}} 
         \\Target: Fridge & Target: Microwave \\
    \end{tabular}
    \caption{Visualization of the path generated by the learned policy. Green is the start point, Red is the end point, and yellow triangles represent the target views.}
    \label{fig:paths}
    \vspace{-0.6cm}
\end{figure}

For comparison, we visualize the paths taken by our agent in similar environments in Fig.~\ref{fig:paths}. It can navigate successfully across a home. The agent takes paths that are straight in narrow passages, but occasionally turns around to explore.

\vspace{-0.1cm}
\section{Conclusion}

We have demonstrated the use of semantic visual representations obtained from state-of-the-art detectors and segmentors for target driven visual navigation. These representations are used as observations in the training of a navigation policy approximated by a deep network. The additional appeal of the proposed choice is the ability to train the navigation policies on synthetic CG data jointly with real images without tackling the domain adaptation problem. 
The detailed ablation studies of different feature representations demonstrate across the board the effectiveness of semantic segmentation and detectors in generalization to previously unseen environments. The effect of adding a RNN component is most 
dominant when synthetic data is added to the training. 
Overall, adding training examples from simulated environments improves the generalization
capability of the proposed approach, except when the depth modality is used with both real and synthetic data. The proposed strategy effectively exploits contextual cues learned from visual representations to guide the agent towards the goal in 54\% of the cases, and outperforms the non-learning based baseline which uses the map and a state-of-the-art detector by 8\%.

\bibliographystyle{splncs}
\bibliography{egbib}

\end{document}